\documentclass{article}

\usepackage{microtype}
\usepackage{graphicx}
\usepackage{booktabs} 

\usepackage{chngcntr}

\usepackage[utf8]{inputenc} 
\usepackage[T1]{fontenc}    
\usepackage{hyperref}       
\usepackage{url}            
\usepackage{booktabs}       
\usepackage{amsfonts}       
\usepackage{nicefrac}       
\usepackage{microtype}      
\usepackage{xcolor}         
\usepackage{fixltx2e}
\usepackage{amsmath}
\usepackage{amsthm}
\usepackage{amssymb}
\usepackage{graphicx}
\usepackage{caption}
\usepackage{enumitem}
\usepackage{adjustbox}
\usepackage{pgfplots}
\usepackage{subcaption}
\usepackage{multirow}

\usepackage{ulem}
\newcommand{\bfs}{\mathbf{s}}

\newcommand{\bff}{\mathbf{f}}

\newcommand{\bfK}{\mathbf{K}}
\newcommand{\bfI}{\mathbf{I}}

\newcommand{\bfA}{\mathbf{A}}
\newcommand{\bfD}{\mathbf{D}}
\newcommand{\bfW}{\mathbf{W}}
\newcommand{\bfS}{\mathbf{S}}

\newcommand{\bfL}{\mathbf{L}}
\newcommand{\bfP}{\mathbf{P}}

\usepackage{hyperref}

\usepackage[accepted]{icml2022}

\icmltitlerunning{pathGCN: Learning General Graph Spatial Operators from Paths}

\begin{document}

\twocolumn[
\icmltitle{pathGCN: Learning General Graph Spatial Operators from Paths}



\icmlsetsymbol{equal}{*}

\begin{icmlauthorlist}
\icmlauthor{Moshe Eliasof}{bgu}
\icmlauthor{Eldad Haber}{ubc}
\icmlauthor{Eran Treister}{bgu}
\end{icmlauthorlist}

\icmlaffiliation{bgu}{Department of Computer Science, Ben-Gurion University, Israel.}
\icmlaffiliation{ubc}{Department of Earth, Ocean and Atmospheric Sciences, University of British Columbia, Canada.}

\icmlcorrespondingauthor{Moshe Eliasof}{eliasof@post.bgu.ac.il}
\icmlcorrespondingauthor{Eran Treister}{erant@cs.bgu.ac.il}

\icmlkeywords{Machine Learning, ICML}

\vskip 0.3in
]



\printAffiliationsAndNotice{} 

\begin{abstract}
Graph Convolutional Networks (GCNs), similarly to Convolutional Neural Networks (CNNs), are typically based on two main operations - spatial and point-wise convolutions.
In the context of GCNs, differently from CNNs, a pre-determined spatial operator based on the graph Laplacian is often chosen, allowing only the point-wise operations to be learnt.
However, learning a meaningful spatial operator is critical for developing more expressive GCNs for improved performance. In this paper we propose pathGCN, a novel approach to learn the spatial operator from random paths on the graph. We analyze the convergence of our method and its difference from existing GCNs. Furthermore, we discuss several options of combining our learnt spatial operator with point-wise convolutions. Our extensive experiments on numerous datasets suggest that by properly learning both the spatial and point-wise convolutions, phenomena like over-smoothing can be inherently avoided, and new state-of-the-art performance is achieved.
\end{abstract}

\section{Introduction}
\begin{figure}[]
    \centering
     \includegraphics[  width=0.5\textwidth]{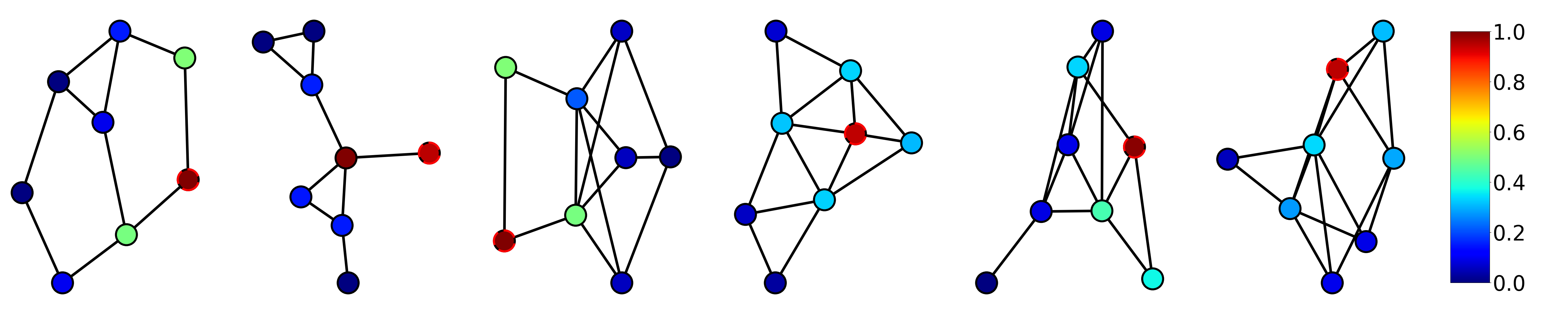}
    \caption{The spatial operator induced by a smoothing kernel on different graphs. The vertex with dashed outline is the path origin. }
    \label{fig:manyGraphsIllustration}
\end{figure}

The study of Graph Convolutional Networks (GCNs) has gained large popularity in recent years \cite{bruna2013spectral, defferrard2016convolutional,
kipf2016semi, bronstein2017geometric, monti2017geometric} in a wide variety of fields and applications such as computer graphics and vision \cite{acnn_boscaini, monti2017geometric, wang2018dynamic, eliasof2020diffgcn}, Bioinformatics \cite{Strokach2020,jumper2021highly}, node classification \cite{kipf2016semi, chen20simple, chamberlain2021grand} and others.
The common ingredient that most of the methods share is the use of a pre-determined \textit{spatial} operator, often times based on the graph Laplacian. While this choice is intuitive and effective, it induces limitations on the behaviour of GCNs. First, it is limited in the aspect of the expressiveness of the networks.  Unlike CNNs \cite{KrizhevskySutskeverHinton2012, he2016deep, howard2017mobilenets}, where both the spatial filters (e.g., $3\times 3$ depth-wise convolutions) and point-wise ($1 \times 1$) convolutions are learnt, here only the latter are left to be determined. Secondly, it is well known \cite{wu2019simplifying} that the Laplacian operator, when applied as in \cite{kipf2016semi} smooths the input features, and therefore a recurrent application of it may lead to over-smoothing, resulting in typically shallow networks. This phenomenon is well documented and studied in the field of GCNs \cite{wu2019simplifying, Zhao2020PairNorm:, chen20simple, chamberlain2021grand, eliasof2021pde}. 
In this work we propose \textit{pathGCN} -- a novel approach that overcomes the limitations above, based on aggregation from random paths defined over the graph vertices.
We show that using this approach it is possible to define spatial operators similarly to the ones used in 2D convolutions on images. Such  operators have variable aperture and coefficients that may increase the expressiveness of GCNs.
In addition, since the coefficients of the operator are learnt, its eigenvalues can be rather different than those of the graph Laplacian. This implies that the learnt kernels can take different roles, from smoothing to edge-detecting (or sharpening) operators. An example of the effective spatial operators that are induced by the smoothing spatial kernel $[0.8, 1.0,0.6]$ is presented in Fig. \ref{fig:manyGraphsIllustration}, where we can see that the spatial operator is dependent both on the graph topology \emph{and} the spatial kernel.

We provide an analysis of our method to motivate our approach in Sec. \ref{sec:method}, and present actual learnt kernels by our network in Fig. \ref{fig:learntKernel}, which suggests that greater spatial flexibility is required to obtain better performance and to avoid over-smoothing, as reflected in our experiments in Sec. \ref{sec:experiments}.

\newpage
Our contributions are as follows:
\begin{itemize}
\item We introduce \textit{pathGCN} -- a novel approach for learning expressive spatial operators for GCNs from random paths. pathGCN supports several formulations, similarly to standard CNNs -- ranging from a global to per layer and per channel learnt spatial operators.
\item We provide an analysis of the behaviour of our pathGCN, and present a stochastic path training policy. 
\item Our experiments reveal the significance of the learnt spatial operator by obtaining and improving the state-of-the-art accuracy on various benchmarks, while also inherently preventing over-smoothing.
\end{itemize}

\section{Related work}
\label{sec:related}

\subsection{Graph convolutional networks}
\paragraph{Notations.} Assume we are given an undirected graph defined as $\cal G=({\cal V},{\cal E})$  where $\cal V$ is a set of $n$ vertices (nodes) and $\cal E$ is a set of $m$ edges. Let us denote by ${\bf f}_i\in\mathbb{R}^c$ the feature vector that resides at the $i$-th node of $\cal G$ with $c$ being the number of channels. Also, we denote the adjacency matrix $\bfA$, where $\bfA_{ij} = 1$ if there exists an edge $(i,j) \in {\cal E}$, and the diagonal degree matrix $\bfD$ where $\bfD_{ii}$ equals to the degree of the $i$-th node. The graph Laplacian is given by $\bfL=\bfD-\bfA$, and its symmetric normalized formulation reads ${\bfL}^{sym} = \bfI - \bfD^{-\frac{1}{2}}\bfA \bfD^{-\frac{1}{2}}$ where $\bfI$ is the identity operator. Let us also denote the adjacency and degree matrices after adding a self-loop to the nodes by $\tilde \bfA$ and $\tilde \bfD$ respectively, and accordingly define the normalized Laplacian of $\mathcal{G}$ with added self-loops by $\tilde \bfL^{sym} = \bfI - \tilde{\bfD}^{-\frac{1}{2}} \tilde{\bfA} \tilde{\bfD}^{-\frac{1}{2}}$. It follows that the spatial operation from GCN \cite{kipf2016semi}, induced by the graph Laplacian is given by $\tilde\bfP = \bfI - \tilde{\bfL}^{sym}$. We refer the readers to \cite{wu2019simplifying} for more information.

\paragraph{The spatial operator in GCNs.}
GCNs typically involve two main ingredients: spatial and point-wise ($1\times1$) convolutions to mix the channels.
The majority of GCNs employ a spatial operator based on the graph Laplacian, followed by a point-wise convolution. For example, GCN \cite{kipf2016semi} is given by:
\begin{equation}
    \label{eq:gcn}
    \bff^{(l+1)} = \sigma(\bfS^{(l)} \bff^{(l)}\bfW^{(l)})
\end{equation}
where $\bfS^{(l)} = \tilde{\bfP} $, and $\bfW^{(l)}$ is a $1 \times 1$ convolution operator.
The combination of the operator $\tilde{\bfP}$ and a learnable $1 \times 1$ convolution continued in a series of works \cite{wu2019simplifying, chen20simple, zhou2021dirichlet}. For instance, the spatial operation in GCNII \cite{chen20simple} can be obtained by replacing $\bfS^{(l)}$ with :
\begin{equation}
\label{eq:gcnii}
\bfS^{(l)} (\bff^{(l)}, \bff^{(0)})=   (1-\alpha^{(l)}) \tilde{\bfP}\bff^{(l)} + \alpha^{(l)}\bff^{(0)},
\end{equation}
with $\alpha^{(l)} \in [0, 1]$ being a hyper-parameter and $\bff^{(0)}$ are the features of the first (embedding) layer, which is also similar to APPNP \cite{klicpera2018combining}.
Another recent example is EGNN \cite{zhou2021dirichlet} which performs
\begin{equation}
\label{eq:dirichlet}
\bfS^{(l)} (\bff^{(l)}, \bff^{(0)}) =   (1-c_{min}) \tilde{\bfP}\bff^{(l)} + \alpha \bff^{(l)} +  \beta\bff^{(0)} ,
\end{equation}
where $c_{min} = \alpha + \beta$, and $\alpha \ , \ \beta$ are learnt scalars.
While the methods above achieve impressive accuracy, they are still limited in their spatial expressiveness due to their dependence on $\tilde{\bfP}$, which may lead to sub-optimal performance.

\begin{figure}
\centering
\begin{tikzpicture}
  \begin{axis}[
      width=1.0\linewidth, 
      height=0.7\linewidth,
      grid=major,
      grid style={dashed,gray!30},
      xlabel=Layers,
      ylabel=Accuracy (\%),
      ylabel near ticks,
      legend style={at={(0.2,0.55)},anchor=north,scale=1.0, draw=none, cells={anchor=west}, fill=none},
      legend columns=1,
      xtick={0,1,2,3,4,5},
      xticklabels = {2,4,8,16,32,64},
      yticklabel style={
        /pgf/number format/fixed,
        /pgf/number format/precision=3
      },
      scaled y ticks=false,
      every axis plot post/.style={thick},
    ]
    \addplot[red, mark=oplus*, forget plot]
    table[x=numLayers,y=path_cora,col sep=comma] {data/path_vs_cheb.csv};
    \addplot[blue, mark=oplus*, forget plot]
    table[x=numLayers,y=path_citeseer,col sep=comma] {data/path_vs_cheb.csv};
    \addplot[red, style=dotted, mark=oplus*, forget plot]
    table[x=numLayers,y=cheby_cora,col sep=comma] {data/path_vs_cheb.csv};
    \addplot[blue, style=dotted, mark=oplus*, forget plot]
    table[x=numLayers,y=cheby_citeseer,col sep=comma] {data/path_vs_cheb.csv};
    \addplot[red, draw=none] coordinates {(1,1)};
    \addplot[blue, draw=none] coordinates {(1,1)};
    \addplot[gray, draw=none] coordinates {(1,1)};
    \addplot[gray, style=dotted, draw=none] coordinates {(1,1)};
    \legend{Cora, Citeseer, pathGCN, ChebNet}
    \end{axis}
\end{tikzpicture}
\caption{ChebNet vs pathGCN on semi-supervised node classification.}
\label{fig:pathVScheb}
\end{figure}
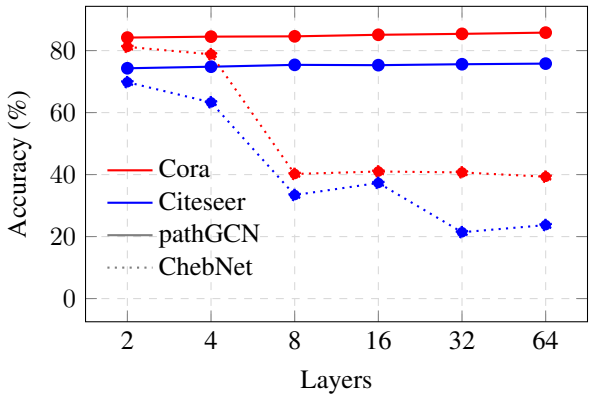

On the other hand, there are methods that allow the freedom of learning a rich spatial operator. For example, ChebNet \cite{defferrard2016convolutional} learns dense convolutions through polynomial parameterization of the graph Laplacian.
While from a conceptual perspective, ChebNet should be able to learn diverse operators and prevent over-smoothing, it still exhibits a degradation in performance when adding more layers, as discussed in \cite{levie2018cayleynets} and according to our experiments as presented in Fig. \ref{fig:pathVScheb}. In addition, GDC \cite{gasteiger_diffusion_2019} propose to impose constraints on the filters of ChebNet to obtain diffusion kernels. Another network to consider is MoNet \cite{monti2017geometric}, which learns local patch operators by a mixture of Gaussian. While the Gaussian parameterization can yield more expressive kernels and favourable results compared to GCN \cite{kipf2016semi}, it is more computationally demanding and challenging to train as more layers are stacked.

\subsection{Random walk on graphs}

The concept of random walk is useful in many applications and domains, from graph node classification \cite{deepWalk,nikolentzos2020random} to RNA disease association \cite{lei2020integrating} and mesh denoising \cite{sun2008random}. In the context of graphs in machine and deep learning, multiple methods utilized random walks for different purposes. DeepWalk \cite{deepWalk} is a two-step algorithm for node embedding learning, based on random walks and the SkipGram \cite{mikolov2013distributed} method. RWGNN \cite{nikolentzos2020random} learns a set of hidden graphs which are then compared with an input graph using a differentiable mutual random walk counting procedure. Our approach is different as we utilize the random walk to effectively sample paths, and employ them to learn a spatial operator by the means of a convolution kernel. Another difference is that RGWNN considers the graph classification problem, as it employs the random walk kernel which is a scalar binary function of two graphs, whose output discards the notion of the graph itself. Therefore, it is not straight-forward to use it for node classification tasks.
The recent PAN \cite{ma2020path} parameterizes the convolution kernel by a polynomial of the adjacency matrix, where the coefficients are the learnt parameters, motivated by path integral theory. However, as its formulation consists of learning non-negative weights, it is prone to over-smoothing and reduced expressiveness. 
Another related work that uses random walk strategy is GraphSAGE \cite{hamilton2017inductive} where neighbourhood sampling of $k$ hops is performed. This method is based on two steps in which an aggregation from two subsequent hops is performed, followed by a $1\times1$ convolution. This is different than ours, as we first extract the complete path, and learn a convolution kernel based on the original information along this path, while the former iteratively aggregates from subsequent hops.

\section{Method}
\label{sec:method}

\begin{figure*}[t]
\centering
    \includegraphics[  width=1\textwidth]{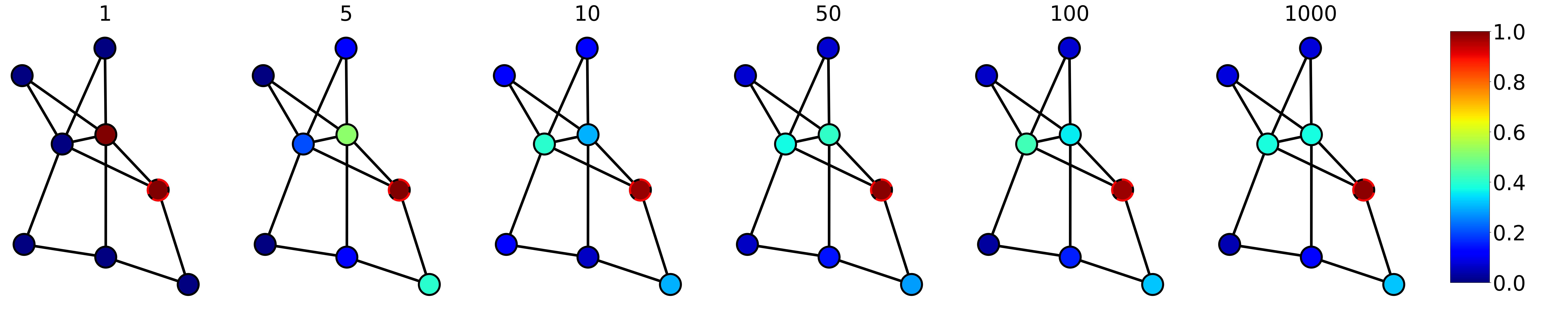}
\caption{The effective spatial operator from a walk of length 3 starting at the dashed node, ranging from 1 to 1000 random walks.}
\label{fig:pathsKernels}
\end{figure*}

\subsection{Learning spatial operators}
\label{sec:learningSpatialOperators}
In this section we motivate the need for learnt spatial operators. To do that, let us first consider a standard CNN where the data resides on a simple uniform mesh-grid.
We note that CNNs and GCNs both represent data with geometrical features. However, while CNNs are networks that operate on a simple mesh-grid graph where pixels (nodes) are linked based on their location, and the local geometry of the graph is fixed, GCNs can be thought of as unstructured meshes where the local geometry varies. 

Given a feature tensor $\bff \in \mathbb{R}^{n \times c}$, 
the convolution in CNNs denoted by $ \bfK\bff $, is a linear operation where each input channel and each output channel has its own spatial operator.
Thus, each linear operator $\bfK$ consists of $c \times c$ different spatial convolutions represented by the tensor $\bfK$.
Furthermore, the convolutions can have a variable aperture (i.e., kernel size) obtaining a larger field of view, and are typically learnt per-layer, yielding a highly expressive set of operators.

In contrast, many popular and recent GCNs \cite{kipf2016semi, wu2019simplifying, chen20simple, chamberlain2021grand, zhou2021dirichlet} and others employ a pre-determined spatial operator, often times guided by the graph Laplacian which is determined solely by the topology of the graph $\mathcal{G}$, coupled with a $1\times 1 $ convolution to mix the channels.
 Therefore, by comparing GCNs to CNNs it is notable that the former have significantly fewer degrees of freedom with respect to their spatial operation. Indeed, 2D CNNs with a spatial kernel of size $k$ and $c$ channels optimize $c \times c \times k \times k$ parameters, while GCNs typically only optimize the $1 \times 1$ convolution, yielding $c \times c$ parameters. In addition to the possible expressiveness issue, the frequent of the graph Laplacian can lead to undesired phenomena such as over-smoothing \cite{wu2019simplifying, Zhao2020PairNorm:, chen20simple}.
Some attempts to overcome the expressiveness limitation consider using polynomials of the graph Laplacian, stabilizing them by constructing a Chebyshev basis \cite{defferrard2016convolutional}. However, it imposes high computational cost due to the frequent computations of the Chebyshev polynomial and fully-connected convolution filters. Furthermore, it is difficult to train such a network due to its eigenvalues distribution, as demonstrated in \cite{levie2018cayleynets}.
With respect to the over-smoothing phenomenon, various techniques were proposed \cite{Zhao2020PairNorm:, chen20simple, Rong2020DropEdge:}. While those methods indeed aid the over-smoothing issue, they do not inherently change the smoothing behaviour of GCNs, but rather ease the smoothing process. 

In what follows, we present a methodology that allows the construction of a graph convolution that is similar to the standard convolution on a regular mesh grid which allows greater expressiveness and inherently does not over-smooth.

\subsection{From fixed to variable spatial operator}

The non-constant topology of the graph is the major obstacle in generating a meaningful spatial convolution in GCNs. We now show that this can be addressed by using random walks.

To this end, we consider a \textit{path} on the graph, on which the weights that parameterize the spatial operator are learnt. Therefore, a transition function that dictates some traversal strategy on the graph is required in order to obtain a path like input to our network.
Specifically, we adopt the graph random walk generator from node2vec \cite{grover2016node2vec}, as implemented in PyTorch-Geometric \cite{pyg2019}, for its simplicity and efficient implementation. We note, however, that a different transition function could also be used or learnt. 

Assume first that we have {\bf a single channel} feature tensor $\bff \in \mathbb{R}^{n \times 1}$, and a path of length $k$ is given. Let us denote the learnt spatial parameters by $\bfs \in \mathbb{R}^k$, and let $y_j = (j_0,...,j_{k-1})$ be a tuple of node indices of a single random path of nodes of length $k$, starting from node $j_0 = j$. The convolution over \emph{a single path} for the $j$-th node is defined by the following linear operator:
\begin{equation}
\label{eq:pathConv_single}
\bfK_{y_j}(\bfs)\bff
= \sum_{i=0}^{k-1}{s_i f_{j_{i}}}.
\end{equation}
That is, the features of the nodes on the path $y_j$ are weighted by the corresponding learnt parameter in $\bfs$, 
and summed to get the feature of the $j$-th node---the node where the path originates from.

More generally, instead of considering a single path, let us sample $p$ different paths, and accordingly define the paths convolution as the average over all sampled paths 
\begin{equation}
\label{pathGCNAv}
 \bfK_{\mathcal{Y}_j}(\bfs)\bff = 
 {\frac 1p} \sum_{y_j \in {\mathcal{Y}_j}} {\bfK_{y_{j}}(\bfs)\bff}  
\end{equation}
where $y_j \in \mathcal{Y}_j$ is a path from a set of $p$ random walks starting from the $j$-th node.

\subsection{Constructing pathGCN}
\label{sec:constructing}
So far we defined the path convolution, which operates in the spatial domain. In what follows, we omit $\bfs$ for brevity, and denote by $\bfK_{\mathcal{Y}}$ the pathConv module described in Eq. \eqref{pathGCNAv} on all the nodes in $\mathcal{V}$ given their corresponding path realizations $\mathcal{Y}$. To obtain a complete network, we need to add the channel mixing convolution and the non-linear activation $\sigma$, as follows
\begin{equation}
    \label{eq:pathGCN_net}
    \bff^{(l+1)} = \sigma(\bfW^{(l)}\bfK_{\mathcal{Y}}^{(l)}\bff^{(l)}) ,
\end{equation}
where in our implementation $\sigma$ is ReLU and $\bfW^{(l)}$ is a $1\times1$ (i.e., point-wise) convolution .
The simplest utilization of pathConv applies a single spatial  operator parameterized by $\bfs \in \mathbb{R}^{k}$, shared among all channels and all $L$ layers. While this is more flexible than using $\tilde{\bfP}$ as in Eq. \eqref{eq:gcn}, it can be further generalized. Scaling up, it is also possible to learn a different parameterization of the spatial operator per layer, that is, $\bfs \in \mathbb{R}^{L \times k}$.
In practice and inspired by modern CNNs \cite{sandler2018mobilenetv2, ephrath2020leanconvnets}, we found that learning a depth-wise spatial operator (i.e., a spatial operator per channel and layer such that $\bfs \in \mathbb{R}^{L \times c \times k}$) followed by a $1 \times 1$ convolution leads to favorable performance both accuracy- and computationally-wise across all the considered data sets in this paper, as reflected in our experiments in Sec. \ref{sec:experiments}. We refer to this architecture as \textit{pathGCN}.
We note that it is also possible to learn different spatial operators for all pairs of channels, per layer -- similarly to a standard fully-connected CNN.

Finally, learning the spatial convolution allows for a variety of operators, from non-smoothing (e.g., edge-detection filters) to smoothing operators. Thus, as we demonstrate in the experiments section, our network achieves state-of-the-art performance, and does not suffer from over-smoothing.

\subsection{Convergence analysis}
\label{sec:convergence}
The process defined in Eq. \eqref{eq:pathConv_single}-\eqref{pathGCNAv} represents a simple stochastic process to build a spatial operation for a single channel that depends on two quantities. First, it depends on the algorithm used to sample the set of paths $\mathcal{Y}$,
and second, it depends on the learnt weights $\bfs$. A natural question that arises is -- does this process converge, and to what? To answer this question we need to make a mild assumption on the random walk process. Namely, we assume that the process is Markovian, that is, at each state the walk can visit any neighbouring node of the current node at an equal probability. In the following, we show that at the sampling limit (i.e., $p \rightarrow \infty$), there exists a stationary distribution of the paths, which induces a spatial operator sampled by our method, as exemplified in Fig. \ref{fig:pathsKernels}. 

Let us consider the simple case of a convolution on a general path of length 2, which can be written as
\begin{equation}\label{pathOne_0}
\bfK_{y_j}(\bfs)\bff =  s_0 f_{j_0} + s_1 f_{j_1}= s_0 f_{j} + s_1 f_{j_1}.
\end{equation}
The second transition follows from the fact that the path originates from the $j$-th node, i.e., $f_{j_0} = f_j$.
Note, that $j_1$ can represent any of the immediate neighbours of the $j$-th node.
Therefore, the expectation of \eqref{pathOne_0} which corresponds to all the nodes $\mathcal{V}$ is given by
\begin{eqnarray}
\label{pathOne}
 {\mathbb E_y} (\bfK_{y}(\bfs)\bff) = s_0 \bff + s_1 (\bfA\bfD^{-1}) \bff 
\end{eqnarray} 
where $\bfA$ is the adjacency matrix and $\bfD$ is a diagonal
matrix with the degree of each node.

More generally, let us consider a path of an arbitrary length $k$.
In this case we have that the expectation term reads
\begin{eqnarray}
\label{eq:pathExpectation}
{\mathbb E_y} \left(\bfK_{y}(\bfs)\bff\right) =  
 \left( \sum_{i=0}^{k-1} s_i  (\bfA \bfD^{-1})^{i} \right) \bff 
\end{eqnarray}
where $(\bfA \bfD^{-1})^{0} = \bfI$.
We note that the transition matrix $\bfA \bfD^{-1}$ is column-stochastic, and therefore its eigenvalues are bounded between $[-1,1]$. Thus, this polynomial formulation is stable and does not diverge for an appropriate choice of coefficients $\bfs$.

Furthermore, Eq. \eqref{eq:pathExpectation} is a deterministic representation of the process given in Eq. \eqref{pathGCNAv}.
If the number of path-realizations, $p$ is large then the results from both are similar. For short paths, a direct evaluation of Eq. \eqref{eq:pathExpectation} can be computationally advantageous compared to its stochastic implementation. However, for long paths, computing the powers of the adjacency matrix times a vector can be expensive. The average over paths is, in this case, an economical way to approximate the process, avoiding repeated matrix multiplications.

Lastly, as we show in our experiments, the stochastic nature of the process has additional advantages. In particular, it allows for better trainability and generalization, compared to the deterministic form. Indeed, a significant increase in the value of $p$ causes performance degradation, as reported in Fig. \ref{fig:ablationFigs}. For inference purposes, we may use both the stochastic or deterministic formulations from Eq. \eqref{pathGCNAv} and \eqref{eq:pathExpectation} respectively, yielding similar results for both.

\subsection{Computational cost and number of parameters}
We compare the cost of our pathGCN with the methods considered in Eq. \eqref{eq:gcn}-\eqref{eq:dirichlet}. On the spatial side, our pathGCN involves $n\times k \times p$ operations rather than $n\times d$ for the considered methods, where $d$ is the mean node degree of the graph. Therefore, if $k \times p$ is larger than $d$, our method requires more computations. Nonetheless, our pathGCN has a wider aperture as it considers paths of $k$ nodes, thus its field of view is larger . We also compare of the number of trainable parameters, which further highlights the difference of our approach. Recall that our pathGCN learns the spatial operator in addition to a $1 \times 1$ convolution, which is also present in other methods. That is, per pathGCN layer, the spatial weights that parameterize $\bfK_{\mathcal{Y}}$ require $c \times k$ parameters, while the $1 \times 1$ convolution $\bfW$ requires $c \times c$ parameters, where typically $k$ is significantly smaller than $c$.
In our experiments, different values of $k$ and $p$ were examined, and we found that setting $k$ between 3 to 7 and $p$ between 5 to 10 achieves better or on par with state-of-the-art models while keeping the computational cost reasonable.

We also consider the stochastic implementation of a random walk over an arbitrary graph. Sampling a single path of length $k$ for all nodes in $\mathcal{V}$ requires $n \times k$ operations\footnote{We note that further efficiency can be gained by using a tree structure that describes the nodes of the sampled paths.}. Therefore, if the number of paths $p$ is smaller than the average node degree $d$ of the graph, then the cost of random walk sampling is smaller than the cost of applying the adjacency matrix as in \eqref{eq:pathExpectation}.
Furthermore, in the deterministic case, a spatial operation on a path of length 2 requires $n \times d$ computations. Extending it to a path of length $k$ requires $n \times d \times (k-1)$ operations, realized by polynomials of degree $k-1$ of the adjacency matrix. We note that Eq. \eqref{eq:pathExpectation}, which is the deterministic form of our method, can also be used for inference purposes.

\section{Experiments}
We demonstrate our pathGCN on node classification and protein-protein interaction \cite{hamilton2017inductive}, followed by an ablation study in order to gain a profound understanding of our method.
In all experiments, we employ a network that is comprised of an embedding layer ($1\times 1$ convolution), followed by a sequence of pathGCN layers, whose final output is fed to a $1 \times 1$ convolution layer which acts as a classifier. A detailed description of the network architecture is given in Appendix \ref{sec:appendix_architecture}. We use the Adam \cite{kingma2014adam} optimizer in all experiments, and perform grid search over the hyper-parameters of our network. The selected hyper-parameters are reported in Appendix \ref{sec:appendix_hyperparams}. The objective function in all experiments is the cross-entropy loss, besides inductive learning on PPI where we use the binary cross-entropy loss. Our code is implemented using PyTorch \cite{pytorch} and PyTorch-Geometric \cite{pyg2019} and trained on an Nvidia Titan RTX GPU.

We show that for all the considered tasks and datasets, whose statistics are provided in Tab. \ref{table:datasets}, our method is either better or on par with other state-of-the-art models.

\begin{table}[]
  \caption{Node classification datasets statistics.}
  \label{table:datasets}
  \begin{center}
  \begin{tabular}{lcccc}
  \toprule
    Dataset & Classes & Nodes & Edges & Features  \\
    \midrule
    Cora & 7 & 2,708 & 5,429 & 1,433 \\
    Citeseer & 6 & 3,327  & 4,732 & 3,703 \\
    Pubmed & 3 & 19,717 & 44,338 & 500 \\
    Chameleon & 5 & 2,277 &  36,101 & 2,325\\
    Cornell & 5 & 183 & 295 & 1,703 \\
    Texas & 5 & 183 & 309 & 1,703 \\
    Wisconsin & 5 & 251 & 499 & 1,703 \\
    PPI & 121 & 56,944 & 818,716 & 50 \\
    Wiki-CS & 10 & 11,701 & 216,123 & 300 \\
    Actor & 5 & 7,600 & 33,544 & 932  \\
    Ogbn-arxiv & 40 & 169,343  & 1,166,243 & 128 \\
    \bottomrule
  \end{tabular}
\end{center}
\end{table}

\label{sec:experiments}
\subsection{Semi-supervised node classification}

\begin{table}[]
  \caption{Summary of semi-supervised node classification accuracy (\%)}
  \label{table:semisupervised_summary}
  \begin{center}
  \begin{tabular}{lccc}
  \toprule
    Method & Cora & Citeseer & Pubmed \\
    \midrule
    ChebNet & 81.2 & 69.8 & 74.4 \\ 
    GCN & 81.1 & 70.8 & 79.0 \\
    GAT  & 83.1 & 70.8 & 78.5 \\
    APPNP & 83.3 & 71.8 & 80.1  \\
    JKNET & 81.1 & 69.8 & 78.1 \\
    GCNII & 85.5 & 73.4 & 80.3 \\
    GRAND & 84.7 & 73.6 & 81.0 \\
    PDE-GCN & 84.3 & 75.6 & 80.6 \\
    EGNN & 85.7 & -- & 80.1 \\
    \midrule
    pathGCN (Ours) &  \textbf{85.8} & \textbf{75.8} & \textbf{82.7} \\
    \bottomrule
  \end{tabular}
\end{center}
\end{table}

Here, we use three datasets -- Cora, Citeseer and Pubmed \cite{sen2008collective}. For all datasets we use the standard training/validation/testing split as in \cite{yang2016revisiting}, with 20 nodes per class for training, 500 validation nodes and 1,000 testing nodes and follow the training scheme of \cite{chen20simple}. For comparison, we consider various models like ChebNet \cite{defferrard2016convolutional}, GCN \cite{kipf2016semi}, GAT\cite{velickovic2018graph}, Inception \cite{szegedy2017inception}, APPNP \cite{klicpera2018combining}, JKNet \cite{jknet}, DropEdge \cite{Rong2020DropEdge:}, GCNII\cite{chen20simple}, GRAND \cite{chamberlain2021grand}, PDE-GCN \cite{eliasof2021pde} and EGNN \cite{zhou2021dirichlet}.

\begin{figure*}
    \centering
    \includegraphics[width=1.0\textwidth]{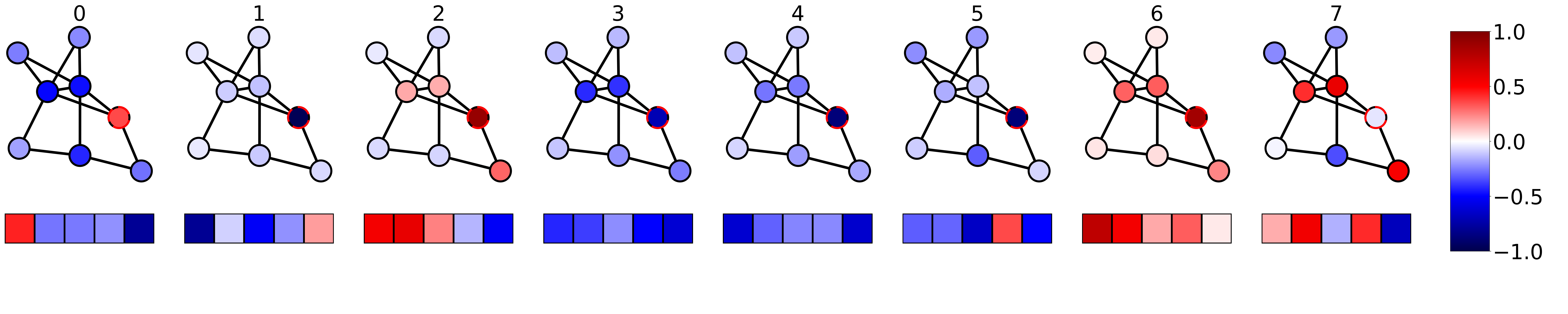}
    \caption{A visualization of the learnt spatial operator $\bfK_{\mathcal{Y}}$ from a 8-layer pathGCN trained on Cora, applied on the dashed node in the graph (top), and its corresponding learnt weights (bottom).
    }
    \label{fig:learntKernel}
\end{figure*}

As discussed in Sec. \ref{sec:method}, our pathGCN is constructed such that wider spatial operators are obtained, allowing for improved expressiveness, namely, compared to methods that are based on the graph Laplacian or its proxies. As objectively portrayed by Tab. \ref{table:semisupervised_summary} and \ref{table:semisupervised}, our pathGCN is capable of obtaining higher accuracy on all three datasets. A bold improvement is obtained on Pubmed, where an accuracy of $82.7\%$ is achieved, compared to previously state-of-the-art GRAND\textsubscript{nl-rw} with $81.0\%$. Our approach also benefits from the inherent absence of over-smoothing, as the spatial operator is fully learnt. This is also validated by inspecting the learnt kernels of different layers of the network. Indeed, as shown in Fig. \ref{fig:learntKernel}, some layers perform smoothing by averaging, while others act as edge-detection filters by considering the difference of neighbouring nodes along the path.

\begin{table}[t]
  \caption{Semi-supervised node classification accuracy ($ \%$). -- indicates not available results.}
  \label{table:semisupervised}
  \begin{center}
  \begin{small}
  \setlength{\tabcolsep}{1.mm}{
  \begin{tabular}{llcccccc}
    \toprule
    \multirow{2}{*}{Dataset} & \multirow{2}{*}{Method} & \multicolumn{6}{c}{Layers} \\
                         &  & 2  & 4  & 8 & 16 & 32 & 64 \\
    \midrule
    Cora & GCN & \bf{81.1} & 80.4 & 69.5 & 64.9 & 60.3 & 28.7 \\
    & GCN (Drop) & \bf{82.8} & 82.0 & 75.8 & 75.7 & 62.5 & 49.5 \\
    & JKNet  & -- & 80.2 & 80.7 & 80.2 & \bf{81.1} & 71.5 \\
    & JKNet (Drop) & -- & \bf{83.3} & 82.6 & 83.0 & 82.5 & 83.2 \\
    & Incep & -- & 77.6 & 76.5 & 81.7 & \bf{81.7} & 80.0  \\
    & Incep (Drop)  & -- & 82.9 & 82.5 & 83.1 & 83.1 & \bf{83.5} \\
    & GCNII  & 82.2 & 82.6 & 84.2 & 84.6 & 85.4 & \bf{85.5} \\
    & GCNII*& 80.2 & 82.3 & 82.8 & 83.5 & 84.9 & \bf{85.3} \\
    & PDE-GCN\textsubscript{D} & 82.0 & 83.6 & 84.0 & 84.2 & 84.3 & \bf{84.3}  \\
    & EGNN & 83.2 & -- & -- & 85.4 & -- & \textbf{85.7}  \\
    & pathGCN (Ours) & 84.2  & 84.5   & 84.6  & 85.1  & 85.4   & \textbf{85.8}  \\
   \midrule
    Citeseer & GCN  & \bf{70.8} & 67.6 & 30.2 & 18.3 & 25.0 & 20.0 \\
    & GCN (Drop) & \bf{72.3} & 70.6 & 61.4 & 57.2 & 41.6 & 34.4 \\
    & JKNet& -- & 68.7 & 67.7 & \bf{69.8} & 68.2 & 63.4 \\
    & JKNet (Drop)  & -- & 72.6 & 71.8 & \bf{72.6} & 70.8 & 72.2 \\
    & Incep  & -- & 69.3 & 68.4 & \bf{70.2} & 68.0 & 67.5 \\
    & Incep (Drop) & -- & \bf{72.7} & 71.4 & 72.5 & 72.6 & 71.0 \\
    & GCNII & 68.2 & 68.8 & 70.6 & 72.9 & \bf{73.4} & 73.4 \\
    & GCNII* & 66.1 & 66.7 & 70.6 & 72.0 & \bf{73.2} & 73.1 \\
    & PDE-GCN\textsubscript{D} & 74.6 & 75.0 & 75.2 & 75.5 & \textbf{75.6} & 75.5 \\
    & pathGCN (Ours) & 74.3  & 74.8   & 75.4 & 75.3 & 75.6 & \textbf{75.8} \\
    \midrule
    Pubmed & GCN & \bf{79.0} & 76.5 & 61.2 & 40.9 & 22.4 & 35.3 \\
    & GCN (Drop) & \bf{79.6} & 79.4 & 78.1 & 78.5 & 77.0 & 61.5 \\
    & JKNet& -- & 78.0 & \bf{78.1} & 72.6 & 72.4 & 74.5 \\
    & JKNet (Drop)  & -- & 78.7 & 78.7 & \bf{79.7} & 79.2 & 78.9 \\
    & Incep & -- & 77.7 & \bf{77.9} & 74.9 & -- & -- \\
    & Incep (Drop) & -- & \bf{79.5} & 78.6 & 79.0 & -- & -- \\
    & GCNII& 78.2 & 78.8 & 79.3 & \bf{80.2} & 79.8 & 79.7 \\
    & GCNII* & 77.7 & 78.2 & 78.8 & \bf{80.3}& 79.8 & 80.1 \\
    & PDE-GCN\textsubscript{D} & 79.3 & \bf{80.6} & 80.1  & 80.4 & 80.2 & 80.3 \\
    & EGNN & 79.2 & -- & -- & 80.0 & -- & \textbf{80.1}  \\
    & pathGCN (Ours) & 81.8 & 81.8 & 82.4 & 82.5 & 82.4 & \textbf{82.7}  \\
    \bottomrule
  \end{tabular}}
  \end{small}
\end{center}
\end{table}

\subsection{Fully-supervised node classification}
To further validate our method, we employ a total of 10 datasets. First, we follow \cite{Pei2020Geom-GCN:} and examine our pathGCN on Cora, Citeseer, Pubmed, Chameleon \cite{musae}, Cornell, Texas and Wisconsin. We also use the same train/validation/test splits of $60 \%, 20\%, 20\%$, respectively, and report the average performance over 10 random splits from \cite{Pei2020Geom-GCN:}. We fix the number of channels to 64 and perform grid search to determine the hyper-parameters. We compare our network with GCN, GAT, Geom-GCN \cite{Pei2020Geom-GCN:}, APPNP, JKNet, Inception, GCNII and PDE-GCN in Tab. \ref{table:fullysupervised}. Our experiments read  improvement across all data-sets compared to all the considered methods. For instance, we obtain $90.02\%$ accuracy on Cora with our pathGCN, compared to $88.49\%$ of GCNII* and $88.60\%$ of PDE-GCN. In addition, we examine our pathGCN on larger datasets using the standard train/validation/test splits of Actor \cite{Pei2020Geom-GCN:}, Ogbn-arxiv \cite{hu2020ogb} and Wiki-CS (20 random splits) \cite{mernyei2020wiki} in Tab. \ref{table:additionalDatasets} and \ref{table:wikiCS} -- where again we see accuracy improvement across all considered datasets.

\begin{table*}[t]
  \caption{Fully-supervised node classification accuracy ($ \%$). (L) denotes the number of layers.}
  \label{table:fullysupervised}
   
  \begin{center}
  \resizebox{2.0\columnwidth}{!}{\begin{tabular}{lccccccc}
    \toprule
    Method & Cora & Cite. & Pubm. & Cham. & Corn. & Texas & Wisc. \\
        \midrule
    GCN \cite{kipf2016semi} & 85.77 & 73.68 & 88.13 &  28.18 &  52.70 & 52.16 & 45.88 \\
    GAT \cite{velickovic2018graph} & 86.37 & 74.32 & 87.62 & 42.93 & 54.32 & 58.38 & 49.41 \\
    Geom-GCN-I \cite{Pei2020Geom-GCN:} & 85.19 & 77.99 & 90.05 & 60.31 & 56.76 & 57.58 & 58.24 \\
    Geom-GCN-P \cite{Pei2020Geom-GCN:} & 84.93 & 75.14 & 88.09 & 60.90 & 60.81  & 67.57 & 64.12 \\
    Geom-GCN-S \cite{Pei2020Geom-GCN:} & 85.27 & 74.71 & 84.75 & 59.96 & 55.68  & 59.73 & 56.67 \\
    APPNP \cite{klicpera2018combining} &  87.87 & 76.53 & 89.40 & 54.30 & 73.51  & 65.41 & 69.02 \\
    JKNet \cite{jknet} & 85.25 (16) & 75.85 (8) & 88.94 (64) & 60.07 (32) & 57.30 (4) & 56.49 (32) & 48.82 (8) \\
    JKNet (Drop) \cite{Rong2020DropEdge:} & 87.46 (16) & 75.96 (8) & 89.45 (64) & 62.08 (32) & 61.08 (4) & 57.30 (32) & 50.59 (8) \\
    Incep (Drop) \cite{Rong2020DropEdge:} & 86.86 (8) & 76.83 (8) & 89.18 (4)  &  61.71 (8) & 61.62 (16) & 57.84 (8) & 50.20 (8) \\
    GCNII \cite{chen20simple} & 88.49 (64) & 77.08 (64) & 89.57 (64) & 60.61 (8)  & 74.86 (16) & 69.46 (32) & 74.12 (16) \\
    GCNII* \cite{chen20simple} & 88.01 (64) & 77.13 (64) & 90.30 (64) & 62.48 (8) & 76.49 (16) & 77.84 (32) & 81.57 (16) \\
    PDE-GCN\textsubscript{M}\cite{eliasof2021pde} & 88.60 (16)  & 78.48 (32)  & 89.93 (16)  & 66.01 (16) & 89.73 (64)  & 93.24 (32)  &  91.76 (16) \\
        \midrule
    pathGCN (Ours) &  \textbf{90.02} (64) &  \textbf{78.95} (32) &  \textbf{90.42} (64) &  \textbf{66.79} (16) &  \textbf{91.35} (8)  &  \textbf{95.14} (16) &  \textbf{93.53} (16) \\   
    \bottomrule
  \end{tabular}}
\end{center}
\end{table*}

\begin{table}[t]
  \caption{Fully-supervised node classification accuracy ($ \%$).}
  \label{table:additionalDatasets}
  \begin{center}
  \resizebox{1.0\columnwidth}{!}{\begin{tabular}{lcc}
    \toprule
    Method & Actor & Ogbn-arxiv \\
        \midrule
    GCN \cite{kipf2016semi}  & 26.86 &  71.74  \\
    GAT \cite{velickovic2018graph} & 28.45 & 71.89   \\
    APPNP \cite{klicpera2018combining}  & 31.26 & 71.82 \\
    Geom-GCN-P \cite{Pei2020Geom-GCN:}  & 31.63 & -- \\
    JKNet \cite{jknet}   & 29.81  & 72.19 \\
    SGC \cite{wu2019simplifying}   & 30.98 & 69.20 \\
    GCNII \cite{chen20simple}  & 32.87 & 72.74 \\
    EGNN \cite{zhou2021dirichlet}  & -- & 72.70  \\ 
    GRAND \cite{chamberlain2021grand}  & --  & 72.23 \\
    \midrule
    pathGCN (Ours) & \textbf{37.54} & \textbf{72.83} \\
    \bottomrule
  \end{tabular}}
\end{center}
\end{table}

\begin{table}[t]
  \caption{Node classification on Wiki-CS.}
  \label{table:wikiCS}
  \begin{center}
  \begin{tabular}{lc}
    \toprule
    Method & Accuracy (\%)\\
        \midrule
    GCN \cite{kipf2016semi} & 77.19  \\
    GAT \cite{velickovic2018graph} & 77.65   \\
    SuperGAT \cite{kim2020findSuperGAT} & 77.90 \\
    APPNP \cite{klicpera2018combining} & 79.84 \\
    \midrule
    pathGCN (Ours) & \textbf{80.02} \\
    \bottomrule
  \end{tabular}
\end{center}
\end{table}

\subsection{Inductive learning}
\label{sec:inductive}
We employ the PPI dataset \cite{hamilton2017inductive} for the inductive learning task. We use a 8 layer pathGCN, without weight-decay, dropout of $0.2$ and a learning rate of 0.001. We compare our results with various methods like GraphSAGE, GAT, JKNet, GeniePath, Cluster-GCN, GCNII and others, and present the micro-averaged F1 score in in Tab. \ref{table:ppi}. We note that our pathGCN achieves a score of $99.61$, superior to methods like GAT, JKNet, GeniePath, PDE-GCN, and slightly above GCNII* with a score of $99.58$.

\begin{table}[t]
  \caption{Inductive learning on PPI dataset. Results are reported in micro-averaged F1 score.}
  \label{table:ppi}
  \begin{center}
  \begin{tabular}{lcc}
    \toprule
    Method & Micro-averaged F1 \\
        \midrule
    GraphSAGE \cite{hamilton2017inductive} & $61.20$ \\
    VR-GCN \cite{vrgcn} & 97.80 \\
    GaAN \cite{zhang18} & 98.71 \\
    GAT \cite{velickovic2018graph} &  97.30 \\
    JKNet \cite{jknet} & 97.60 \\
    GeniePath \cite{geniepath} & 98.50 \\
    Cluster-GCN \cite{clustergcn} & 99.36 \\
    GCNII \cite{chen20simple} & 99.54 \\
    GCNII* \cite{chen20simple} & 99.58 \\
    PDE-GCN\textsubscript{M} \cite{eliasof2021pde} & 99.18  \\
    \midrule
    pathGCN (ours) & \textbf{99.61} \\
    \bottomrule
  \end{tabular}
\end{center}
\end{table}

\subsection{Graph classification}
\label{sec:graphClassification_experiment}
Our previous experiments considered various datasets and settings of the node classification task. To further demonstrate the efficacy of our pathGCN we experiment with graph classification on the popular TUDatasets \cite{Morris2020TUDatasets}. We follow the same experimental settings from \cite{xu2018how} and report the 10 fold cross-validation performance on MUTAG, PTC, PROTEINS and NCI1 datasets. The hyper-parameters are determined by grid search, as in \cite{xu2018how} and are reported in Appendix \ref{sec:appendix_hyperparams}. We compare our pathGCN with recent methods like DGCNN \cite{zhang2018end}, IGN \cite{maron2018invariant},GIN \cite{xu2018how}, CIN \cite{bodnar2021CW} and GSN \cite{bouritsas2022improving}.
The results are summarized in Tab. \ref{table:graphClassification}, where our pathGCN shows better or similar results compared to the considered methods, further highlighting the efficacy of our approach.

\begin{table}
  \caption{TUDatasets graph classification accuracy (\%).}
  \label{table:graphClassification}
  \begin{center}
  \resizebox{1.00\columnwidth}{!}
  {\begin{tabular}{lcccc}
  \toprule
    Model & MUTAG & PTC & PROTEINS & NCI1 \\
    \midrule
    DGCNN & $85.8_{\pm1.8}$ & $58.6_{\pm2.5}$ & $75.5_{\pm0.9}$ & $74.4_{\pm0.5}$ \\
    IGN & $83.9_{\pm13.0}$ & $58.5_{\pm6.9}$ & $76.6_{\pm5.5}$ & $74.3_{\pm2.7}$   \\
    GIN & $89.4_{\pm5.6}$ & $64.6_{\pm7.0}$ & $76.2_{\pm2.8}$ & $82.7_{\pm1.7}$ \\
    CIN & $92.7_{\pm6.1}$ & $68.2_{\pm5.6}$ & $77.0_{\pm4.3}$ & $\bold{83.6_{\pm1.4}}$  \\
    GSN & $92.7_{\pm7.5}$ & $68.2_{\pm7.2}$ & $76.6_{\pm5.0}$ & $83.5_{\pm2.0}$ \\
    \midrule
    pathGCN (ours) & $\bold{94.7_{\pm4.7}}$ & $\bold{75.2_{\pm5.3}}$ &  $\bold{80.4_{\pm4.2}}$ & $83.3_{\pm1.3}$ \\
    \bottomrule
  \end{tabular}}
\end{center}
\end{table}

\subsection{Inference and runtimes}

\begin{table}[t]
  \caption{Deterministic vs stochastic pathGCN inference accuracy (\%).}
  \label{table:compare_inference}
  \begin{center}
  \resizebox{1.0\columnwidth}{!}
  {\begin{tabular}{lcccccc}
  \toprule
    Inference & Cora & Cite.  & Pub. & ogbn-arxiv & Wisc. & PPI \\
    \midrule
    Determin. & 85.8 & 75.8 & 82.7 & 72.84 & 93.51 & 99.61  \\
        \multirow{2}{*}{Stochastic}  & 85.8 & 75.8 & 82.7 & 72.83 & 93.53 & 99.61 \\
        & ${\pm0.29}$ & ${\pm0.34}$ & ${\pm0.34}$ & ${\pm0.13}$ & ${\pm0.21}$ & ${\pm0.02}$ \\
    \bottomrule
  \end{tabular}}
\end{center}
\end{table}

\begin{table}[t]
  \caption{Computation times (in ms) on Cora.}
  \label{table:compare_runtimes}
  \begin{center}
  \resizebox{1.00\columnwidth}{!}
  {\begin{tabular}{lcccc}
  \toprule
    Model & Path samp. & Training  & Inference & Acc (\%) \\
      \midrule
    GCN (2 layers) &- & 4.07 & 1.97 & 81.1 \\
    GCNII (2 layers) &- & 4.24 & 1.95 & 82.2 \\
    GCNII (8 layers) &- & 13.05 & 6.87 & 84.2 \\
    GAT (2 layers) & -&5.27 & 1.96 & 83.1 \\
    \midrule
    pathGCN$_{k=2, p=5}$ (2 layers) & 0.378 & 3.48 
   & 1.67 & 81.3   \\
   pathGCN$_{k=5, p=5}$ (2 layers) & 0.384 & 5.35 & 2.17 & 84.2 \\
    \bottomrule
  \end{tabular}}
\end{center}
\end{table}

 In this section we compare the inference accuracy of our stochastic approach with its deterministic form. Later, we report the runtimes of our method.
 
As discussed in Sec. \ref{sec:convergence}, both the stochastic and deterministic forms of our pathGCN can be used during inference. In Tab. \ref{table:compare_inference}, we show that after training our stochastic pathGCN, it is possible to obtain similar results (with up to $0.1\%$ accuracy difference) by its deterministic form as in Eq. \eqref{eq:pathExpectation}. The stochastic results are averaged over 10 inference runs, and we also present their standard deviation.

Next, we present the training and inferences times, as well as path sampling time of our pathGCN and compare it with GCN, GCNII and GAT, in Tab. \ref{table:compare_runtimes}. For reference, we also report the node classification accuracy of the different methods. We see that the sampling time is relatively small, and that our pathGCN has a similar runtime to GCN, GCNII and GAT, while obtaining similar or better accuracy. For example, our pathGCN with $k=5, p=5$ obtains an accuracy of $84.2\%$ on Cora and requires 5.35 milliseconds (ms) for a training iteration, while GCNII requires 8 layers and 13.05 ms to achieve the same accuracy.

\subsection{Ablation  study}

\begin{figure*}
\centering
\begin{subfigure}[t]{.48\linewidth}
    \centering
    \begin{tikzpicture}
      \begin{axis}[
          width=1.0\linewidth, 
          height=0.5\linewidth,
          grid=major,
          grid style={dashed,gray!30},
          xlabel=p,
          ylabel=Accuracy (\%),
          ylabel near ticks,
          legend style={at={(0.45,0.4)},anchor=north,scale=0.8, draw=none, cells={anchor=west}, font=\tiny, fill=none},
          legend columns=1,
          xtick={0,1,2,3,4,5,6},
          xticklabels = {1,2,5,10,20,50,100},
          yticklabel style={
            /pgf/number format/fixed,
            /pgf/number format/precision=3
          },
          scaled y ticks=false,
          every axis plot post/.style={thick},
        ]
        \addplot
        table[x=numpaths,y=cora,col sep=comma] {data/numPaths_ablation.csv};
        \addplot
        table[x=numpaths,y=citeseer,col sep=comma] {data/numPaths_ablation.csv};
        \addplot
        table[x=numpaths,y=pubmed,col sep=comma] {data/numPaths_ablation.csv};
        \end{axis}
    \end{tikzpicture}
    \end{subfigure}
    \begin{subfigure}[t]{.48\linewidth}
    \centering
    \begin{tikzpicture}
      \begin{axis}[
          width=1.0\linewidth, 
          height=0.5\linewidth,
          grid=major,
          grid style={dashed,gray!30},
          xlabel=k,
          ylabel=Accuracy (\%),
          ylabel near ticks,
          legend style={at={(0.8,0.55)},anchor=north,scale=1.0, draw=none, cells={anchor=west}, font=\small, fill=none},
          legend columns=1,
          xtick={0,1,2,3,4,5,6},
          xticklabels = {1,2,3,5,7,9,11},
          yticklabel style={
            /pgf/number format/fixed,
            /pgf/number format/precision=3
          },
          scaled y ticks=false,
          every axis plot post/.style={thick},
        ]
        \addplot
        table[x=kernelSize,y=cora,col sep=comma] {data/kernelSize_ablation.csv};
        \addplot
        table[x=kernelSize,y=citeseer,col sep=comma] {data/kernelSize_ablation.csv};
        \addplot
        table[x=kernelSize,y=pubmed,col sep=comma] {data/kernelSize_ablation.csv};
        \legend{Cora, Citeseer, Pubmed}
        \end{axis}
    \end{tikzpicture}
    \end{subfigure}%
\caption{Semi-supervised node classification accuracy (\%), as a function of number of paths $p$ and path length $k$.}
\label{fig:ablationFigs}
\end{figure*}
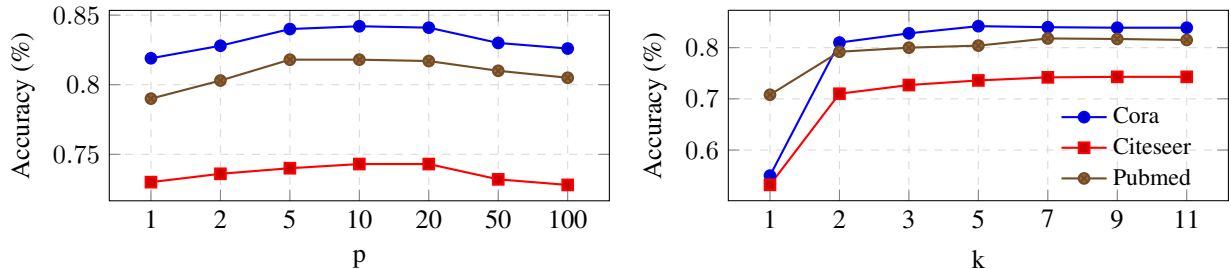

In this section we consider the different possible variants of our pathGCN, and the influence of our its hyper-parameters the path length $k$ and the number of paths $p$. 
As discussed in Sec. \ref{sec:method}, our method can be formulated in various manners. That is, it is possible to learn a \textit{global} spatial operator, which is most similar to GCN \cite{kipf2016semi}, only there it is fixed and not learnt. We denote this variant by pathGCN\textsubscript{G}. In addition, we can learn a spatial operator \textit{per layer} that is shared among the channels, denoted by pathGCN\textsubscript{PL}. The next step is our pathGCN from Sec. \ref{sec:constructing}, which is more similar to CNNs, by utilizing a \textit{depth-wise} spatial operator (i.e., per channel and layer).
As depicted in Tab. \ref{table:ablationVariants}, the global variant, pathGCN\textsubscript{G} yields the least attractive performance among the three considered variants. This is not surprising, as a global operator is learnt, which is less expressive than pathGCN\textsubscript{PL} and pathGCN. Still, it is interesting to see that no over-smoothing is evident. As discussed in \cite{wu2019simplifying}, a recurrent application of a \textit{fixed} operator leads to over-smoothing. However, the learnt spatial operator of pathGCN\textsubscript{G} is variable.
Following that, the per layer variant, pathGCN\textsubscript{PL} prevents over-smoothing and also further improves accuracy. This is obtainable as the network has the freedom to learn a variety of kernels, which may function as smoothing kernels or edge detectors. 
Finally, we see that the additional degrees of freedom in the depth-wise operator in pathGCN lead to overall better performance, also as depicted in our other experiments.

As for the hyper-parameters of our method, we report in Fig. \ref{fig:ablationFigs} the influence of the number of paths $p$ and path length $k$ on the performance of our pathGCN on semi-supervised node classification using Cora, Citeseer and Pubmed. In all the experiments we use a 2 layer pathGCN, fix one hyper-parameter, and vary the other. Specifically, for the evaluation of the influence of number of paths $p$, we fix the path length to $k=5$, and the results in Fig. \ref{fig:ablationFigs} suggest that indeed the stochastic nature of our method is beneficial to obtain higher accuracy, since we see a consistent accuracy degradation trend as $k$ is increased past 20.
In addition, we present in Fig. \ref{fig:ablationFigs}, that by fixing $p=10$ and examining a variable path length $k$ from 1 to 11, accuracy improves and stagnates at $k=7$. We can see that when $k=1$, our pathGCN behaves similarly to an MLP \cite{qi2017pointnet} (as it considers only the self node), and that in some cases, increasing the kernel size caused a slight performance degradation.

\begin{table}[t]
  \caption{Variants of pathGCN on semi-supervised classification. Results are reported in accuracy ($ \%$).}
  \label{table:ablationVariants}
  \begin{center}
  \begin{small}
  \setlength{\tabcolsep}{1.mm}{
  \begin{tabular}{llcccccc}
    \toprule
    \multirow{2}{*}{Dataset} & \multirow{2}{*}{Method} & \multicolumn{6}{c}{Layers} \\
                         &  & 2  & 4  & 8 & 16 & 32 & 64 \\
    \midrule
    Cora &  pathGCN\textsubscript{G} & 83.0 & 81.9  & 81.5  & 81.2  & 81.4 & 82.0    \\
    & pathGCN\textsubscript{PL} & 82.9 & 83.3 & 83.5 & 83.8 & 84.1 & 84.7 \\
    & pathGCN & 84.2  & 84.5   & 84.6  & 85.1  & 85.4   & 85.8   \\
   \midrule
    Citeseer &  pathGCN\textsubscript{G} &  73.1  & 71.9  & 72.0 & 71.9  & 72.6  & 71.7   \\
    & pathGCN\textsubscript{PL} & 73.4  & 73.6  &  74.0 &  74.3 & 74.5  & 75.0   \\
    & pathGCN & 74.3  & 74.8   & 75.4 & 75.3 & 75.6 & 75.8 \\
    \midrule
        Pubmed & pathGCN\textsubscript{G} & 80.9   & 80.4  & 81.1 & 81.0 & 80.4   &  80.8  \\
    & pathGCN\textsubscript{PL} & 81.1  & 81.2  & 81.5  &  82.0 & 82.2  & 82.1   \\
    & pathGCN  & 81.8 & 81.8 & 82.4 & 82.5 & 82.4 & 82.7  \\

    \bottomrule
  \end{tabular}}
  \end{small}
\end{center}
\end{table}

\section{Conclusion}
In this paper we propose a new approach for learning the spatial operators for GCNs. Our motivation stems from the need for deep GCNs that have expressive spatial kernels, similar to standard CNNs that do not over-smooth. Our approach leverages on paths defined on the graph, to enable the learning of such operators, further bridging the gap between GCNs and CNNs. 

Just as the Laplacian is not the sole spatial operator used on images in CNNs, it may also not necessarily be optimal in the case of graphs and GCNs. To this end we propose \textit{pathGCN} which replaces the Laplacian based operator by a fully learnt kernel. Indeed, our experiments reveal that more expressive kernels can be learnt based on the data and task at hand, leading to consistently better accuracy on numerous applications and datasets and without over-smoothing.

\section*{Acknowledgements}
The research reported in this paper was supported by the Israel Innovation Authority through Avatar consortium. In addition, this work was supported in part by the Israeli Council for Higher Education (CHE) via the Data Science Research Center, Ben-Gurion University of the Negev, Israel. ME is supported by Kreitman High-tech scholarship.
\bibliography{references}
\bibliographystyle{icml2021}

\clearpage
\appendix
\section{Architecture in details}
\label{sec:appendix_architecture}
We elaborate on the architecture that was used in our experiments. As discussed in Sec. \ref{sec:experiments}, our network is comprised of an embedding layer ($1
\times 1$ convolution), a sequence of pathGCN layers, and a closing (projection) layer ($1
\times 1$ convolution).
Throughout this section, $c_{in}$ and $c_{out}$ denote the input and output channels, respectively, and $c$ denotes the number of features in hidden layers (which is a hyper-parameter, as given in Appendix \ref{sec:appendix_hyperparams}) We denote the number of pathGCN blocks by $L$, and the dropout probability by $p_{drop}$.
Our architecture for node classification tasks is described in Tab. \ref{table:basicArch}, which is similar to the architecture found in GCNII \cite{chen20simple}, only with our pathGCN instead. In Tab. \ref{table:graphClassificationArch} we present our network for graph classification tasks, which is based on the one in GIN \cite{xu2018how}.

\begin{table}[ht!]
  \caption{pathGCN architecture for node classification.}
  \label{table:basicArch}
  \begin{center}
  \begin{tabular}{lcc}
  \toprule
    Input size & Layer  &  Output size \\
    \midrule
    $n \times c_{in}$ & $1\times1$ Dropout($p_{drop}$) & $n \times c_{in}$ \\
    $n \times c_{in}$ & $1\times1$ Convolution & $n \times c$ \\
    $n \times c$ & ReLU & $n \times c$ \\    
    $n \times c$ & $L \times $ pathGCN block & $n \times c$ \\
    $n \times c$ & $1\times1$ Dropout($p_{drop}$) & $n \times c$ \\
    $n \times c$ & $1\times1$ Convolution & $n \times c_{out}$ \\
    \bottomrule
  \end{tabular}
\end{center}
\end{table}

\begin{table}[h]
  \caption{pathGCN architecture for graph classification.}
  \label{table:graphClassificationArch}
  \begin{center}
  \begin{tabular}{lcc}
  \toprule
    Input size & Layer  &  Output size \\
    \midrule
    $n \times c_{in}$ & $1\times1$ Convolution & $n \times c$ \\
    $n \times c$ & ReLU & $n \times c$ \\    
    $n \times c$ & $L  \times [ $ pathGCN , BN  & $n \times c$   \\
    &  $, 1\times1$ Convolution, ReLU $]$ &  \\
    $n \times c$ & $1\times1$ Add-pool & $1 \times c$ \\
    $1 \times c$ & $1\times1$ Convolution & $1 \times c$ \\

    $1 \times c$ & $1\times1$ Dropout($p_{drop}$) & $1 \times c$ \\
    $1 \times c$ & $1\times1$ Convolution & $1 \times c_{out}$ \\
    \bottomrule
  \end{tabular}
\end{center}
\end{table}


\section{Hyper-parameters details}
\label{sec:appendix_hyperparams}
We provide the selected hyper-parameters in our experiments. We denote the learning rate of our pathGCN layers by $LR_{GCN}$, and the learning rate of the $1\times 1$ opening (embedding) and closing (classifier) layers by $LR_{oc}$.
Also, the weight decay for the opening and closing layers is denoted by $WD_{oc}$, and for the pathGCN layers by $WD_{GCN}$.

\subsection{Semi-supervised node classification}
The hyper-parameters for this experiment are summarized in Tab. \ref{table:semisupervisedHyperParams}.

\begin{table}
  \caption{Semi-supervised node classification hyper-parameters}
  \label{table:semisupervisedHyperParams}
  \begin{center}
  \resizebox{1.0\columnwidth}{!}{\begin{tabular}{lccccccccc}
  \toprule
    Dataset & $LR_{GCN}$ &  $WD_{GCN}$ & $LR_{oc}$ & $WD_{oc}$ &  $c$ & $p_{drop}$ &  $k$  & $p$ \\
    \midrule
    Cora & $1\cdot 10^{-3}$ & $2 \cdot 10^{-5}$ & $1 \cdot 10^{-2}$ & $1\cdot 10^{-5}$  &  $64$ & $0.6$ & $5$ &  $5$  \\
    \midrule
    Citeseer & $1\cdot 10^{-3}$ & $ 1\cdot 10^{-5}$ &  $7 \cdot 10^{-3}$ & $5\cdot 10^{-5}$ & $256$ &  $0.7$ & $5$ & $5$  \\
    \midrule
    Pubmed & $5\cdot 10^{-3}$ & $0$ & $1 \cdot 10^{-2}$ & $1\cdot 10^{-5}$  &$256$ & $0.5$ & $7$ & $10$ \\
    \bottomrule
  \end{tabular}}
\end{center}
\end{table}

\subsection{Fully-supervised node classification}
The hyper-parameters for this experiment are summarized in Tab. \ref{table:fullysupervisedHyperParameters}.

\begin{table}[t]
  \caption{Fully-supervised node classification hyper-parameters}
  \label{table:fullysupervisedHyperParameters}
  \begin{center}
  \resizebox{1.0\columnwidth}{!}{\begin{tabular}{lccccccccc}
  \toprule
    Dataset & $LR_{GCN}$ &  $WD_{GCN}$  & $LR_{oc}$ & $WD_{oc}$ &  $c$ & $p_{drop}$ &  $k$ & $p$  \\
    \midrule
    Cora & $1\cdot 10^{-4}$ & $1\cdot 10^{-4}$ & $7 \cdot 10^{-2}$ & $1 \cdot 10^{-4}$ & $64$ & $0.5$ & $5$ & $10$     \\
    \midrule
    Citeseer & $3 \cdot 10^{-4}$ & $5\cdot 10^{-5}$ & $8 \cdot 10^{-3}$ & $1 \cdot 10^{-4}$ & $64$ & $0.5$ & $5$ & $10$ \\
    \midrule
    Pubmed & $1\cdot 10^{-4}$ & $2\cdot 10^{-4}$  & $1 \cdot 10^{-2}$ & $1 \cdot 10^{-6}$ & $64$ & $0.5$& $7$ & $10$  \\
       \midrule
    Chameleon & $5 \cdot 10^{-4}$ & $1\cdot 10 ^{-5}$ & $5 \cdot 10^{-3}$ & $3 \cdot 10^{-5}$ & $64$ & $0.5$ & $3$ & $10$  \\
       \midrule
    Cornell & $4 \cdot 10^{-4}$ & $1\cdot 10^{-5}$ & $5 \cdot 10^{-2}$ & $5 \cdot 10^{-4}$ & $64$ & $0.5$ & $5$ & $10$ \\
       \midrule
    Texas & $3 \cdot 10^{-4}$ & $5 \cdot 10^{-4}$  &$4 \cdot 10^{-2}$ & $1\cdot 10^{-4}$ & $64$ & $0.5$ & $7$ & $10$\\
       \midrule
    Wisconsin & $3 \cdot 10^{-4}$ & $2 \cdot 10^{-4}$  & $1 \cdot 10^{-2}$ & $5\cdot 10^{-5}$ & $64$ & $0.5$ & $7$ & $10$ \\
           \midrule
    Actor & $2\cdot 10^{-4}$ & $1\cdot 10^{-4}$ & $8 \cdot 10^{-2}$ & $5\cdot 10^{-4}$ & $64$ & $0.5$ & $7$ & $10$ \\ 
           \midrule
    Wiki-CS & $3 \cdot 10^{-2}$  & $1\cdot 10^{-4}$  & $7 \cdot 10^{-3}$ & $1\cdot 10^{-5}$ & $64$ & $0.3$ & $7$ & $5$ \\
           \midrule
    Ogbn-arxiv  & $1 \cdot 10^{-3}$ & $0$ & $1 \cdot 10^{-3}$ & $0$ & $256$ & $0.1$ & $5$ & $10$ \\    
    \bottomrule
  \end{tabular}}
\end{center}
\end{table}

\subsection{Inductive learning on PPI}
For this experiment we used $LR_{oc} = LR_{GCN} = 0.001$ , with dropout probability $p_{drop} = 0.2$, $k=5$ and $p=10$, and no weight decay was used.

\subsection{Graph classification}
The hyper-parameters in this experiment were chosen according to the protocol described in \cite{xu2018how}. We present the chosen hyper-parameters in Tab. \ref{table:graphClassificationHyperParams}. Throughout all the experiments in this section, no weight decay was used.

\begin{table}[ht!]
  \caption{Graph classification hyper-parameters. BS denotes batch size.}
  \label{table:graphClassificationHyperParams}
  \begin{center}
  \resizebox{1.0\columnwidth}{!}{\begin{tabular}{lcccccccccc}
  \toprule
     Dataset & $LR_{GCN}$ & $LR_{oc}$  &   $c$ & $p_{drop}$ & BS & p & k \\
    \midrule
   MUTAG & 0.01 & 0.01  & 32 & 0 & 32 & 5 & 5 \\
    PTC &  0.01 & 0.01  & 32 & 0 & 32 & 5 & 5  \\
    PROTEINS &  0.01  & 0.01  & 32 & 0 & 128  & 5 & 10\\
    NCI1 &  0.01  & 0.01   & 32 & 0.5 & 32 & 5 & 10\\
    \bottomrule
  \end{tabular}}
\end{center}
\end{table}

\subsection{Ablation study}
In this experiment we used the same hyper-parameters as reported in Tab. \ref{table:semisupervisedHyperParams}, for the results in Tab. \ref{table:ablationVariants}. For the results in Fig. \ref{fig:ablationFigs}, we use the same learning rate and weight decay, but $k$ and $p$ are as described in the main paper.

\end{document}